# ASMFS: Adaptive-Similarity-based Multi-modality Feature Selection for Classification of Alzheimer's Disease

Yuang Shi[1], Chen Zu[2], Mei Hong[1], Luping Zhou[3], Lei Wang[4], Xi Wu[5], Jiliu Zhou[1,5], Daoqiang Zhang[6], Yan Wang[1,*]

[1] School of Computer Science, Sichuan University, Chengdu, China.
[2] Risk Controlling Research Department, JD.com, Chengdu, China.
[3] School of Electrical and Information Engineering, University of Sydney, Australia.
[4] School of Computing and Information Technology, University of Wollongong, Australia
[5] School of Computer Science, Chengdu University of Information Technology, Chengdu, China.
[6] School of Computer Science, Nanjing University of Aeronautics and Astronautics, Nanjing, China.

**Abstract:** Multimodal classification methods using different modalities of imaging and non-imaging data have great advantages over traditional single-modality-based ones for the diagnosis and prognosis of Alzheimer's disease (AD), as well as mild cognitive impairment (MCI) which is the prodromal stage of AD. With the increasing amount of high-dimensional heterogeneous data to be processed, multi-modality feature selection has become a crucial research direction in medical image analysis. However, traditional methods usually depict the data structure using fixed and predefined similarity matrix as a priori, which is difficult to precisely measure the intrinsic relationship structure across different modalities in high-dimensional spaces. In addition, based on the predefined similarity matrix, the chosen neighbors are suboptimal thus limiting the performance of the subsequent classification task. To overcome these drawbacks, in this paper, we propose a novel multi-modal feature selection method called Adaptive-Similarity-based Multi-modality Feature Selection (ASMFS) which performs adaptive similarity learning and feature selection simultaneously. Specifically, a similarity matrix is learned by jointly considering different modalities and at the same time, to obtain the associated subspace representation, an efficient feature selection is conducted by imposing group sparsity-inducing $\ell_{2,1}$-norm constraint. The regularization parameter within the similarity optimization problem is set as a variable to be learned for reducing the number of parameters. Furthermore, an effective optimization algorithm is designed to solve the proposed algorithm. To validate our method, we perform experiments on the Alzheimer's Disease Neuroimaging Initiative (ADNI) database using baseline MRI and FDG-PET imaging data collected from 51 AD, 43 MCI converters (MCI-C), 51 MCI non-converters (MCI-NC) and 52 normal controls (NC). According to the experimental results, our proposed method captures the intrinsic similarity across multiple modalities with the change of feature dimension which in turn helps to select the most instructive features and ultimately promotes the performance of classification. The empirical study also well demonstrates that the proposed joint learning method outperforms the existing state-of-the-art approaches for multi-modality classification of AD/MCI.

---

[*]Corresponding Author.

# 1. Introduction

Alzheimer's disease (AD) is a chronic neurodegenerative disease which is the main cause of dementia, leading to problems with language, disorientation, mood swings, bodily functions and, ultimately, death [1]. According to a recent report by Alzheimer's Association, about 5.3 million Americans have AD and 5.1 million are old people who are aged over 65 [2]. In 2050, the AD population will increase to beyond 100 million [3]. Although some therapies may temporarily improve symptoms of AD, there is no treatment that stops or reverses its progression so far. Hence, the early diagnosis of AD and especially its prodromal condition known as mild cognitive impairment (MCI) is highly essential for timely therapy. For the last decades, neuroimaging technique has proven to be a powerful tool to investigate the characteristics of neurodegenerative progression between AD and normal controls (NC), for instance, structural MR imaging (MRI) for brain atrophy measurement [4-7], functional imaging (e.g., FDG-PET) for hypometabolism quantification [8, 9], and cerebrospinal fluid (CSF) for quantification of specific proteins [6, 10-12].

In recent years, machine learning and pattern classification methods have been widely applied for the early diagnosis of AD based on single modality of biomarkers. For example, Lei et al. [13] propose to build a framework based on longitudinal multiple time points data to predict clinical scores of AD. Liu et al. [14] developed an inherent structure-based multi-view learning method which utilizes the structure information of MRI data well. In addition to structural MRI, some researchers also used fluorodeoxyglucose positron emission tomography (FDG-PET) for AD or MCI classification [15-17]. However, since such single-task learning treats each task as a stand-alone one without considering the intrinsic association among different tasks, the performance of these approaches is only suboptimal in predicting the progression of brain diseases.

As is known, different modalities of biomarkers can provide the inherently complementary information. For example, structural MRI reveals patterns of gray matter atrophy, while FDG-PET measures the reduced glucose metabolism in the brain. It is reported that MRI and FDG-PET provide different sensitivity for memory prediction between disease and health [18]. As a result, many studies have used multimodal data to further improve classification performance. For instance, Tong et al. [19] present a multi-modality classification framework using nonlinear graph fusion to efficiently exploit the complementarity in the multi-modal data of PET and CSF. Hinrichs et al. [20] combined two modalities, i.e., MRI and PET, for classification of AD. Zhang et al. [21] combined three modalities, i.e., MRI, FDG-PET and CSF, to classify AD/MCI from NC. Gray et al. [22] used MRI, FDG-PET, CSF and categorical genetic information for AD/MCI classification. These existing studies have suggested that different imaging modalities provide different views of brain structure or function that could be overlooked by using single modality. Thus, utilizing modalities together to improve the accuracy in disease diagnosis becomes sensible idea for researches.

Despite the promising performance of the above multi-modality classification methods, they all face the challenge of handling high dimensionally features for the analysis. The curse of dimensionality, that occurs when there are insufficient training subjects versus large feature dimensions, limits the further performance improvement of existing methods. In addition, the high dimensional feature vectors usually contain some irrelevant and redundant features and thus lead to the overfitting problem, which hurts the generalization ability of the algorithm. Moreover, in neuroimaging data analysis, features may correspond to brain regions. In such a case, feature selection can detect the regions with brain atrophy, pathological amyloid depositions or metabolic alterations, thus becomes potentially useful for timely therapy of brain diseases. Therefore, feature selection is a compelling topic for multi-modality data in medical image analysis.

However, there are two main challenges of multi-modality feature selection [23]. First, because feature representations extracted from different modalities may have distinct distributions in a variety of feature spaces, it is challenging to integrate these discriminative features into a unified form of feature representation. Second, since various features from different modalities play distinctive roles in classification task, how to evaluate each feature group and select the relevant features for the task remains a problem. Concentrating on the above challenges, several multi-modality feature selection methods have been developed in recent years. For example, a multi-task feature selection (MTFS) was proposed in [24] to select common subset of relevant features from each modality. Liu et al. [25] proposed a multi-task feature selection method (i.e., inter- modality multi-task feature selection (IMTFS)) to preserve the complementary inter-modality information. Different from MTFS, IMTFS imposes an inter-modality term, which can maintain the geometry structure of different modalities from the same subject. Also, a manifold regularized multi-task feature learning method (M2TFS) [26] was proposed to preserve the data distribution information in each modality separately. In these approaches, MTFS focuses on feature selection, without considering the underlying data structure. IMTFS and M2TFS not only focus on feature selection, but also take into account of the relationship of training subjects.

To preserve the underlying data structure, similarity measure plays an important role. It is worth noting that, in the above mentioned multi-modality feature selection methods, the neighbors and the similarity between the original high-dimensional data are usually obtained separately from each individual modality, such as the case in M2TFS. Besides, the similarity is fixed before feature selection in above methods. However, considering the existence of noisy and redundant features, the relationship of subjects in high-dimensional space may not necessarily reveal the underlying data structure in the low-dimensional space after feature selection. On the other hand, the performance of a mass of machine learning methods, such as K-Nearest Neighbors (KNN) [27], Support Vector Machine (SVM) [28], RBF Neural Network [29] as well as K-means [30] and other clustering algorithms, to a large extent, are determined by the similarity between each pair of the subjects. Thus, the performance of above methods which are dependent on subject similarity would degrade if the similarity matrix is constructed inaccurately.

Therefore, this paper argues that the similarity should not be fixed but adaptive to change with the low-dimensional representation after feature selection. In other words, the similarity is supposed to be variable and optimized while selecting multi-modality features. Besides, it is commonly accepted that a large amount of real-world high-dimensional data actually lie on low-dimensional manifolds embedded within a high-dimensional space [31]. Provided there is sufficient data (such that the manifold is well-sampled), we expect each data point and its neighbors to lie on or close to a locally linear patch of the manifold. We can characterize the local geometry of these patches by linear coefficients that are used to reconstruct each data point from its neighbors [32]. Additionally, since neighborhood similarity is more reliable compared with the similarity retrieved from farther samples, preserving local neighborhood structure is of great help to construct an accurate similarity matrix. Therefore, instead of updating the similarities between every data pairs, we only consider the local neighborhood similarities of every subject.

In light of this, we propose a novel learning method which is able to simultaneously capture the intrinsic similarity shared across different modality data, and select the most informative features. This method is called Adaptive-Similarity-based Multi-modality Feature Selection (ASMFS). Specifically, the proposed method includes two major steps: 1) adaptive similarity learning as well as multi-modality feature selection, and 2) multimodal classification. Our contributions are in Step One. In step one, we first simultaneously update the similarity matrix and optimize the sparse regression coefficient which performs feature selection. Furthermore, the manifold hypothesis discussed above is introduced in adaptive similarity learning and $K$ local neighborhood similarities of every subject will be updated at one time, where *K* is the number of similarities which will calculated for each subject. Moreover, to better depict the collective information among multiple modalities, we assume that the similarity matrix is shared by different modality data collected from the same subject. The proposed objective function is optimized in an alternating manner. Then, in step two, we use multi-kernel support vector machine (SVM) to fuse the selected features from multi-modality data for the final classification. Our method is evaluated for AD classification, where we use the baseline MRI and FDG-PET data from the ADNI database including 51 AD, 43 MCI-C, 51 MCI-NC and 52 NC. Experimental comparison with the existing methods on the ADNI database illustrates that the proposed method not only yields improved performance on identifying disease status, but also discovers the disease sensitive biomarkers.

The rest of this paper is organized as follows. Section 2 introduces our proposed multi-modality feature selection architecture and methodology. Experiments and experimental results are presented in Section 3. Finally, we discuss and conclude this paper in Section 4 and Section 5.

## 2. Method

### 2.1 Multi-modality feature selection with adaptive similarity learning

In this section, we first introduce how to learn similarity measure from both single and multi-modality data through adaptive similarity learning. Then, we further show how to embed this similarity learning into our multi-modality feature selection framework. The selected features are eventually taken in a multi-kernel support vector machine for disease classification [33]. Figure 1 gives the overview of the proposed classification method.

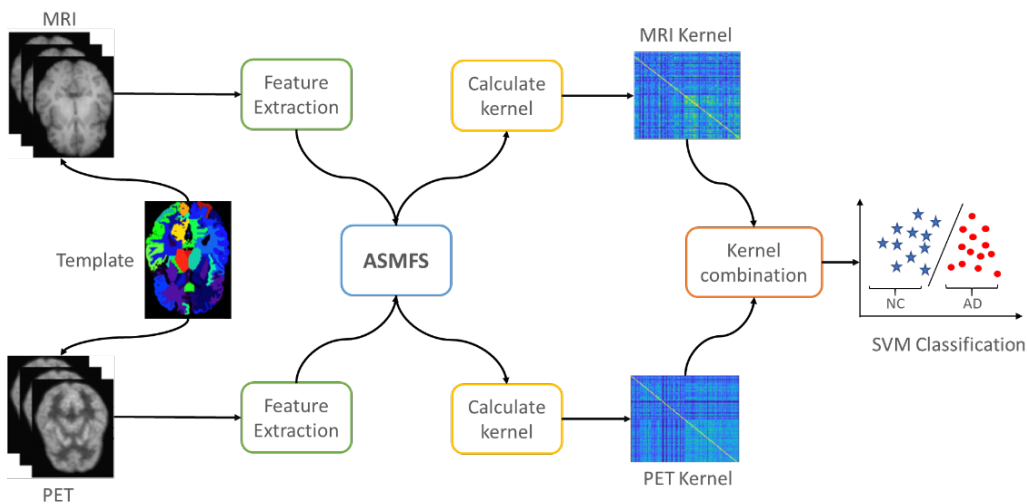

Fig. 1 Framework of multi-modality feature selection with adaptive similarity learning

2.1.1 Adaptive similarity learning

Let's consider the single modality scenario first. Suppose that in a $d$ dimensional feature space, the data matrix of $n$ subjects is denoted as $X = [x_1, x_2, \ldots, x_n] \in \mathbb{R}^{d \times n}$. The subjects can be divided into $c$ classes and the corresponding label vector is given as $y = [y_1, y_2, \ldots, y_n]$. The similarity matrix $S$ that indicates the similarity of data pairs can be constructed by two assumptions: 1) it is hoped that the similarity between $x_i$ and $x_j$ can be reflected by their Euclidean distance. If the distance $||x_i - x_j||_2^2$ between $x_i$ and $x_j$ is small, the similarity $s_{ij}$ should be large, 2) if $x_i$ and $x_j$ belong to different classes, the similarity $s_{ij}$ should be zero. In this section, we will first discuss two ideal cases according to above two assumptions, and then propose our case with consideration of the two cases.

To begin with, we formulate the following objective to determine the similarities $s_{ij}$ based on aforementioned assumptions:

$$\begin{aligned}
\min_{s_i} \quad & \sum_{j=1}^{n} \|x_i - x_j\|_2^2 \, s_{ij}, \\
\text{s.t.} \quad & s_i^T \mathbf{1} = 1, 0 \leq s_{ij} \leq 1, \\
& s_{ij} = 0, \text{ if } y_i \neq y_j,
\end{aligned} \quad (1)$$

where $s_i \in \mathbb{R}^n$ is a vector of which the $j$-th entry is $s_{ij}$ and $\mathbf{1}$ denotes a column vector with all the elements as one. However, by solving problem (1), it can be found that only one which is the closest neighbor to $x_i$ has the similarity $s_{ij} = 1$, while the others are 0. In other words, it is a trivial solution.

Then, suppose the distance information is unavailable between subjects and the following problem is solved to estimate the similarities:

$$\begin{aligned}
\min_{s_i} \quad & \sum_{j=1}^{n} s_{ij}^2, \\
\text{s.t} \quad & s_i^T \mathbf{1} = 1, 0 \leq s_{ij} \leq 1, \\
& s_{ij} = 0, \text{ if } y_i \neq y_j.
\end{aligned} \quad (2)$$

The solution of $s_{ij} = \frac{1}{n}$ reveals that all the subjects will become the nearest neighbors of $x_i$ with $\frac{1}{n}$ probability. The problem (2) can be actually regarded as the prior of the nearest neighbor probability when the pairwise subject distance is unknown. Considering problem (1) and (2) jointly, we solve the following objective to obtain the similarities $s_{ij}$:

$$\min_{\mathbf{s}_i} \sum_{j=1}^{n} \left( \|\mathbf{x}_i - \mathbf{x}_j\|_2^2 s_{ij} + \alpha s_{ij}^2 \right),$$
$$\text{s.t} \quad \mathbf{s}_i^T \mathbf{1} = 1, 0 \leq s_{ij} \leq 1, \quad (3)$$
$$s_{ij} = 0, \text{ if } y_i \neq y_j.$$

The second term $s_{ij}^2$ can be regarded as a regularization term to avoid the trivial solution in problem (1) and $\alpha$ is the regularization parameter. The problem (3) can be applied to calculate the similarities for each subject $\mathbf{x}_i$. Consequently, in this paper we estimate the similarities for all subjects by solving the following problem:

$$\min_{\forall i, \mathbf{s}_i} \sum_{i=1}^{n} \sum_{j=1}^{n} \left( \|\mathbf{x}_i - \mathbf{x}_j\|_2^2 s_{ij} + \alpha s_{ij}^2 \right),$$
$$\text{s.t} \quad \mathbf{s}_i^T \mathbf{1} = 1, 0 \leq s_{ij} \leq 1, \quad (4)$$
$$s_{ij} = 0, \text{ if } y_i \neq y_j.$$

We can transform problem (4) to linearly constrained quadratic programming which can be solved by KKT conditions. And the matrix $\mathbf{S} = [\mathbf{s}_1, \mathbf{s}_2, \ldots, \mathbf{s}_n]^T \in \mathbb{R}^{n \times n}$ can be treated as a similarity matrix of $n$ subjects.

Now, we extend the above adaptive similarity learning to multi-modality case. The multi-modality data are denoted as $\mathbf{X}_1, \mathbf{X}_2, \ldots, \mathbf{X}_M$, where $M$ is the number of modalities. The data matrix of the $m$-th modality is defined as $\mathbf{X}_m = [\mathbf{x}_1^{(m)}, \mathbf{x}_2^{(m)}, \ldots, \mathbf{x}_n^{(m)}]$. For all the multi-modality data, we solve the following problem to obtain the similarity matrix $\mathbf{S}$:

$$\min_{\mathbf{S}} \sum_{i=1}^{n} \sum_{j=1}^{n} \left( \sum_{m=1}^{M} \|\mathbf{x}_i^{(m)} - \mathbf{x}_j^{(m)}\|_2^2 s_{ij} + \alpha s_{ij}^2 \right),$$
$$\text{s.t} \quad \mathbf{s}_i^T \mathbf{1} = 1, 0 \leq s_{ij} \leq 1, \quad (5)$$
$$s_{ij} = 0, \text{ if } y_i \neq y_j.$$

Please note that different from traditional multi-modality methods which calculate the similarity for each modality separately, the similarity matrix $\mathbf{S}$ obtained in (5) is shared by different modality data. By doing so, we can not only assume that the different modality data collected from the same subject should be generated via the same intrinsic distribution, but also capture the information collectively from multiple modalities. Thus the similarities of these data in diverse modalities would be identical.

Then, we embed the adaptive similarity learning into multi-modality feature selection in order to learn the optimal neighborhood similarity for feature selection, thereby improving the performance of multi-modality classification by utilizing the more discriminative information.

### 2.1.2 Multi-modality feature selection with adaptive similarity learning

To integrate the similarity learning problem (5) with multi-modality feature selection, the objective function of our proposed method is defined as:

$$\min_{W,S} \sum_{m=1}^{M} \|y - w_m^T X_m\|_2^2 + \mu \|W\|_{2,1}$$
$$+ \lambda \sum_{i=1}^{n} \sum_{j=1}^{n} \left( \sum_{m=1}^{M} \|w_m^T x_i^{(m)} - w_m^T x_j^{(m)}\|_2^2 s_{ij} + \gamma s_{ij}^2 \right), \quad (6)$$
$$\text{s.t} \quad s_i^T \mathbf{1} = 1, 0 \le s_{ij} \le 1,$$
$$s_{ij} = 0, \text{ if } y_i \ne y_j.$$

where $W = [w_1, w_2, \dots, w_M] \in \mathbb{R}^{d \times M}$ is the coefficient matrix, $w_m \in \mathbb{R}^d$ is the coefficient of the $m$-th modality. The $\ell_{2,1}$ norm of $W$ is defined as $\|W\|_{2,1} = \Sigma_i^d \sqrt{\Sigma_j^M w_{ij}^2}$, which can result in sparse rows of $W$ to achieve feature selection. ASMFS considers different modalities of subjects into similarity construction. $\lambda$, $\mu$ and $\gamma$ are regularization parameters to balance the terms in (6).

From (6), we can not only capture the inherent similarity shared across different modality data, but also select the most informative features.

## 2.2 Optimization algorithm

The objective function (6) is optimized in an alternate manner. Specifically, we fix $W$ and optimize $S$ and then fix $S$ and optimize $W$.

**1) Fix $W$ and optimize $S$.**

Removing the irrelative part to $S$ from (6), we can get the following objective:

$$\min_S \sum_{i=1}^{n} \sum_{k \in \{k|y_i=y_k\}} \left( \sum_{m=1}^{M} \|w_m^T x_i^m - w_m^T x_k^m\|_2^2 s_{ik} + \gamma s_{ik}^2 \right),$$
$$\text{s.t.} \quad \sum_k^n s_{ik} = 1, \quad (7)$$
$$0 \le s_{ij} \le 1.$$

In Section 2.1.1, we assumed that if $x_i$ and $x_j$ belong to different classes, the similarity $s_{ij}$ should be zero. So when $k \in \{k|y_i \ne y_k\}$, then $s_{ik} = 0$, which means $\|w_m^T x_i^m - w_m^T x_k^m\|_2^2 s_{ik} = 0$. Hence, we only need to consider the similarity between subjects from the same class, i.e., $s_{ik}$ when $k \in \{k|y_i = y_k\}$.

Since the similarity learning of one subject is independent with respect to the learning of the others, we can safely decompose the similarity of individual subject according to the objective from (7):

$$\min_{s_i} \Sigma_{k\in\{k|y_i=y_k\}} \left(\Sigma_{m=1}^{M}\|w_m^T x_i^m - w_m^T x_k^m\|_2^2 \, s_{ik} + \gamma s_{ik}^2\right),$$
$$\text{s.t. } \Sigma_k^n s_{ik} = 1, \quad (8)$$
$$0 \leq s_{ij} \leq 1.$$

By defining $d_{ik} = \Sigma_{m=1}^{M} ||w_m^T x_i^m - w_m^T x_k^m||_2^2$, (8) can be simplified to the following form:

$$\min_{s_i} \Sigma_{k\in\{k|y_i=y_k\}}(d_{ik}s_{ik} + \gamma_i s_{ik}^2)$$
$$= \min_{s_i} \Sigma_{k\in\{k|y_i=y_k\}} \left(\gamma_i \left(s_{ik} + \frac{1}{2\gamma_i}d_{ik}\right)^2 - \frac{d_{ik}^2}{4\gamma_i}\right)$$
$$= \min_{s_i} \Sigma_{k\in\{k|y_i=y_k\}} \left(\gamma_i \left(s_{ik} + \frac{1}{2\gamma_i}d_{ik}\right)^2\right) \quad (9)$$
$$= \min_{s_i} \gamma_i \Sigma_{k\in\{k|y_i=y_k\}} \left(s_{ik} + \frac{1}{2\gamma_i}d_{ik}\right)^2$$
$$= \min_{s_i} \gamma_i \left\|s_i + \frac{1}{2\gamma_i}d_i\right\|_2^2.$$

Accordingly, the following objective is formulated:

$$\min_{s_i} \left\|s_i + \frac{1}{2\gamma_i}d_i\right\|_2^2,$$
$$\text{s.t.} \Sigma_{k=1}^{n} s_{ik} = 1, \quad (10)$$
$$0 \leq s_{ij} \leq 1.$$

We can solve (10) by KKT conditions.

The above objective is a convex function which can be solved utilizing Lagrange method:

$$L(s_i, \eta, \beta) = \frac{1}{2}\left\|s_i + \frac{1}{2\gamma_i}d_i\right\|_2^2 - \eta(s_i^T \mathbf{1} - 1) - \beta^T s_i, \quad (11)$$

where $\beta \geq 0$, $\eta \geq 0$ are Lagrange multipliers and $\mathbf{1}$ denotes a column vector with all the elements as one. Taking the derivative of (11) with respect to $s_i$ and setting it equal to 0, we have:

$$\frac{\partial L(s_i,\eta,\beta)}{\partial s_i} = \frac{\partial}{\partial s_i}\left(\frac{1}{2}\left\|s_i + \frac{1}{2\gamma_i}d_i\right\|_2^2 - \eta(s_i^T \mathbf{1} - 1) - \beta^T s_i\right) = 0. \quad (12)$$

The optimal solution can be figured out by KKT [34] condition:

$$s_{ik} = \left(-\frac{d_{ik}}{2\gamma_i} + \eta\right)_+ = \max\left(-\frac{d_{ik}}{2\gamma_i} + \eta, 0\right). \quad (13)$$

Practically, as discussed in Section 1, keeping the local manifold structure of data is proved well effective [35, 36] in feature selection. One can improve the classification performance with attention only to the local structure of data. Therefore, we expect to learn a sparse $s_i$. That is, only the nearest $K$ neighbors of $x_i$ have the opportunity to connect with $x_i$. Moreover, sparse similarity matrix learning is of great help to reduce the computational burden for the later processing.

Let us suppose that $d_{i1}, d_{i2}, \ldots, d_{in}$ are sorted from the lowest to the highest. Provided that the optimal $s_i$ has only $K$ none-zero elements, using (13), we know $s_{iK} > 0$ and $s_{i,K+1} \leq 0$. Hence, the following inequalities hold:

$$\begin{cases} s_{ik} = -\frac{d_{ik}}{2\gamma_i} + \eta > 0, k \leq K \\ s_{ik} = -\frac{d_{ik}}{2\gamma_i} + \eta \leq 0, k > K. \end{cases} \tag{14}$$

Substituting the constraint $\sum_{k=1}^{K} s_{ik} = 1$ into the (14), we have:

$$\sum_{k=1}^{K}\left(-\frac{d_{ik}}{2\gamma_i} + \eta\right) = 1$$
$$\Rightarrow \eta = \frac{1}{K} + \frac{1}{2K\gamma_i}\sum_{k=1}^{K} d_{ik}. \tag{15}$$

Plugging $\eta$ into (14) leads to the constraint of $\gamma_i$:

$$\frac{K}{2} d_{iK} - \frac{1}{2}\sum_{k=1}^{K} d_{ik} < \gamma_i \leq \frac{K}{2} d_{i,K+1} - \frac{1}{2}\sum_{k=1}^{K} d_{ik}. \tag{16}$$

Finally, we have the optimal $\gamma_i$:

$$\gamma_i = \frac{K}{2} d_{i,K+1} - \frac{1}{2}\sum_{k=1}^{K} d_{ik}. \tag{17}$$

**2) Fix S and optimize W.**

Removing the irrelative part to $W$ from (6), we get the following objective:

$$\min_{W} \mathcal{L}(W) = \sum_{m=1}^{M}\sum_{i}^{n}\|y_i - w_m^T x_i^m\|_2^2 + \mu \|W\|_{2,1}$$
$$+ \lambda \sum_{i=1}^{n}\sum_{k \in \{k|y_i=y_k\}}\left(\sum_{m=1}^{M}\|w_m^T x_i^m - w_m^T x_k^m\|_2^2 s_{ik}\right). \tag{18}$$

Inspired by [37], we solve (18) using the weighted and iterative method. When the row elements in $W$ are nonezero, that is $w_{i,:} \neq 0, i = 1,2,\ldots,d$, we take the derivative of $\|W\|_{2,1}$:

$$\frac{\partial \|W\|_{2,1}}{\partial W} = 2DW, \tag{19}$$

where $D \in \mathbb{R}^{d \times d}$ is a diagonal matrix, the $i$-th diagonal element is:

$$d_{ii} = \frac{1}{2} ||w_{i,:}||_2^{-1}. \tag{20}$$

When $D$ is fixed, taking the derivative of $W$ in (18) is equivalent to doing so in the following objective:

$$\min_W \mathcal{L}(W) = \sum_{m=1}^M \sum_i^n ||y_i - w_m^T x_i^m||_2^2 + \lambda Tr(W^T D W)$$

$$+ \mu \sum_i^n \sum_{k \in \{k|y_i=y_k\}} \left( \sum_{m=1}^M ||w_m^T x_i^m - w_m^T x_k^m||_2^2 s_{ik} \right). \tag{21}$$

Please note that, the analytical form of $W$ can be obtained via solving (21), and therefore (21) substitutes (18) in our learning framework. The procedure is summarized in Algorithm 1.

---
**Algorithm 1**: Multi-modality feature selection with adaptive similarity learning
---
**Input**: Multi-modality sample matrix $\{X^1, X^2, ..., X^M\}$ and label matrix $y = [y_1, y_2, ..., y_n]$.
**Initial**: $\lambda > 0, \mu > 0, K > 0, D = I.$
**Repeat**
1. Update $W$ using Eq. (21);
2. Update $D$ using Eq. (20);
3. Update $S$ using Eq. (13);
**Until converges**
**Output**: $W$
---

## 2.3 Multi-kernel support vector machine

Multi-kernel support vector machine (MKSVM) [38] is adopted for classification after feature selection processing. First, we generate a kernel matrix $k^m(x_i^m, x_j^m) = \phi^m((x_i^m)^T(x_j^m))$ for each modality data after feature selection. Then, the $M$ kernel matrices are linearly combined $k(x_i, x_j) = \sum_{m=1}^M \beta_m k^m(x_i^m, x_j^m)$ before fed to MKSVM for training, where $\sum_{m=1}^M \beta_m = 1, \beta_m \geq 0$. It is notable that in our experiments, the optimal $\beta_m$ is determined via a coarse-grid search through cross-validation on the training set. Finally, when an unseen subject comes, the trained MKSVM model is able to predict the category of the new subject by the following decision function:

$$f(x) = \text{sign}(\sum_{i=1}^n y_i \alpha_i \sum_{m=1}^M \beta_m k^m(x_i^m, x^m) + b). \tag{22}$$

## 3. Experiments

### 3.1 Dataset and settings

*Dataset*: The data involved in this paper are obtained from the Alzheimer's Disease Neuroimaging Initiative (ADNI) database (www.loni.usc.edu). The ADNI was launched by a wide range of academic institutions and private corporations and the subjects were collected from approximately 200 cognitively normal older individuals to be followed for 3 years, 400 MCI patients to be followed for 3 years, and 200 early AD patients to be followed for 2 years across the United States and Canada.

We use imaging data from 202 ADNI participants with corresponding baseline MRI and PET data. In particular, it includes 51 AD patients, 99 MCI patients and 52 NC. The MCI patients were divided into 43 MCI converters (MCI-C) who have progressed to AD with 18 months and 56 MCI non-converters (MCI-NC) whose diagnoses have still remained stable within 18 months. Table 1 lists the clinical and demographic information for the study population.

Table 1 Subject information

|  | AD | MCI-C | MCI-NC | NC |
|---|---|---|---|---|
| Subjects number | 51 | 43 | 56 | 52 |
| Age | 75.2±7.4 | 75.8±6.8 | 74.7±7.7 | 75.3±5.2 |
| Education | 14.7±3.6 | 16.1±2.6 | 16.1±3.0 | 15.8±3.2 |
| MMSE | 23.8±2.0 | 26.6±1.7 | 27.5±1.5 | 29.0±1.2 |
| CDR | 0.7±0.3 | 0.5±0.0 | 0.5±0.0 | 0.0±0.0 |

In this study, image pre-processing is performed for all MRI and PET images following the same procedures as in [39-41]. Specifically, N3 algorithm [42] is employed to correct the intensity inhomogeneity after anterior commissure-posterior commissure correlation performed. For MRI data, gray matter (GM) is segmented by FAST [43] and then the GM tissue volume of each region obtained according to a 93 manual labels template is chosen as a feature. After the alignment to the respective MRI image, the average intensity of each ROI in the PET image is calculated as a feature. Therefore, there are totally 93 features for MR image and 93 features for PET image.

***Comparison methods:*** In order to assess the classification performance, the proposed method is compared with six existing multimodal classification methods including 1) multi-kernel SVM [38] (denoted as MKSVM), 2) multi-kernel method with LASSO [56] feature selection performed independently on single modalities (denoted as lassoMKSVM), 3) multi-kernel method using multi-modal feature selection method (denoted as MTFS) proposed in [44], and 4) multi-kernel method with manifold regularized multitask feature learning (denoted as M2TFS) proposed in [45]. We also 5) concatenate 93 features from MRI images and 93 features from PET images into a 186-dimension vector, and then utilize LASSO as feature selection method followed by the standard SVM with linear kernel for classification (denoted as lassoSVM). Besides, 6) we use the standard SVM with linear kernel as a contrast method.

***Validation:*** Z-score normalization $f'_i = (f_i - \overline{f_i})/\sigma_i$ is performed on every feature $f_i$ from MRI images and PET images separately, where $\overline{f}_i$ and $\sigma_i$ respectively represent the mean and the standard deviation of the $i$-th feature from the training set. It is noted that we performed Z-score normalization on test set with $\overline{f}_i$ and $\sigma_i$ calculated from the training set.

Performance measurements including accuracy (ACC), sensitivity (SEN), specificity (SPE), area under receiver operating characteristic (ROC) curve (AUC) and F1 Score are utilized in the experiments to quantify the classification performance of different methods. The 10-fold cross-validation strategy is adopted due to the limited subjects. Specifically, the whole set of subject samples are equally partitioned into 10 subsets, from which 9 subsets were randomly selected for training and the remaining subset for testing. The above procedure was repeated 10 times to avoid any bias caused by the partition.

***Hyper-parameters***: In our method, there are three hyper-parameters, i.e., the sparsity regularizer $\lambda$, the adaptive similarity learning regularization term $\mu$, and the number of neighbors $K$. The above parameters are determined by another 10-fold cross-validation. $\lambda$, μ and $K$ are searched in the range {0.1,5,20,60,100}, {0,5,10,15,20} and {1,3,5,7,9}, respectively. Moreover, in multi-kernel SVM with a linear kernel, $C$ is set as 1 and the kernel combination coefficients $\beta_{MRI}$, $\beta_{PET}$ are chosen from 0.1 to 1.0 with step 0.1 and constrained with $\beta_{MRI}+\beta_{PET}$=1and $\beta_{MRI}$, $\beta_{PET}$≥0.

### 3.2 Classification Results
### 3.2.1 AD vs. NC classification

Table 2 lists the classification results of AD vs. NC produced by seven methods. The standard deviations are given and the best results are denoted in bold in Table 2. As observed, our proposed method (ASMFS) consistently achieves the best performance from all the methods in comparison. Specifically, ASMFS achieves the accuracy of 96.76%, the sensitivity of 96.1%, the specificity of 97.47%, the AUC of 0.9703 and the F1 Score of 96.63. Among the three multi-modality feature selection methods, i.e., MTFS, M2TFS and ASMFS, we can find that over 95% accuracy is achieved by M2TFS and ASMFS, indicating that maintaining inter-modality information is effective for feature selection. Furthermore, compared with MTFS and M2TFS which adopt fixed similarity, the better performance achieved by ASMFS indicates that adaptive similarity is of great help to depict more accurate data distribution after feature selection. The lowest accuracy of 88.24% achieved by SVM is because SVM does not perform feature selection but directly uses the raw feature vectors for classification. Moreover, the multi-modality features are simply concatenated without utilizing the complementary information from different modalities. lassoSVM achieves better performance than SVM because it adopts LASSO as feature selection which removes redundant features and noise. Comparing the results between MKSVM and SVM, it can be found that utilizing complementary information from different modalities greatly promotes the accuracy (3%), which indicates the necessity of jointly considering multiple modalities for the diagnosis of AD. On the other side, MKSVM overlooks the redundant features. Thus, after feature selection by LASSO, lassoMKSVM gets better accuracy than MKSVM. This also verifies that redundancy exists in the original data.

Table 2 Comparison of performance of different methods for AD vs. NC classification

| Method | Accuracy (%) | Sensitivity (%) | Specificity (%) | F1 Score | AUC |
| --- | --- | --- | --- | --- | --- |
| SVM | 88.24±0.0972 | 91.07±0.1155 | 85.57±0.1591 | 88.61±0.0925 | 0.9471±0.0007 |
| lassoSVM | 90.90±0.0873 | 90.60±0.1240 | 91.23±0.1233 | 90.71±0.0900 | 0.9460±0.0007 |
| MKSVM | 91.87±0.0875 | 92.30±0.1249 | 91.63±0.1160 | 91.68±0.0927 | 0.9526±0.0007 |
| lassoMKSVM | 92.33±0.0739 | 93.47±0.1030 | 91.30±0.1261 | 92.41±0.0726 | 0.9534±0.0007 |
| MTFS | 92.52±0.0816 | 93.77±0.1115 | 91.37±0.1213 | 92.50±0.0846 | 0.9541±0.0007 |
| M2TFS | 95.00±0.0707 | 94.67±0.1009 | 95.40±0.0826 | 94.85±0.0740 | 0.9636±0.0006 |
| **ASMFS** | **96.76±0.0545** | **96.10±0.0836** | **97.47±0.0660** | **96.63±0.0573** | **0.9703±0.0006** |

Sensitivity measures the proportion of patients that are correctly diagnosed, while specificity measures the percentage of healthy people who are correctly classified. Obviously, the best score for sensitivity and specificity is 100%, but in practice, it is nearly impossible to achieve such performance. As for medical diagnosis, identifying patients as healthy people is worse than identifying healthy people as patients for the reason that the latter is only needed more examinations. Therefore, specificity is considered more important than sensitivity. As found in Table 2, both high sensitivity and specificity are achieved by our proposed method, indicating that our method rarely overlooks an AD patient or misclassifies a normal individual as the diseased. Furthermore, excepting lassoSVM, M2TFS and ASMFS, the sensitivity of other methods is higher than the specificity.

We can see from Table 2 that ASMFS not only achieves the best performance, but also keeps the lowest standard deviation, indicating the proposed method is more stable. Figure 2 plots the corresponding ROC curves of above methods for AD vs. NC classification, from which we can see the proposed method obtains the best performance, with high TPR at low FPR, and larger AUC value than other methods.

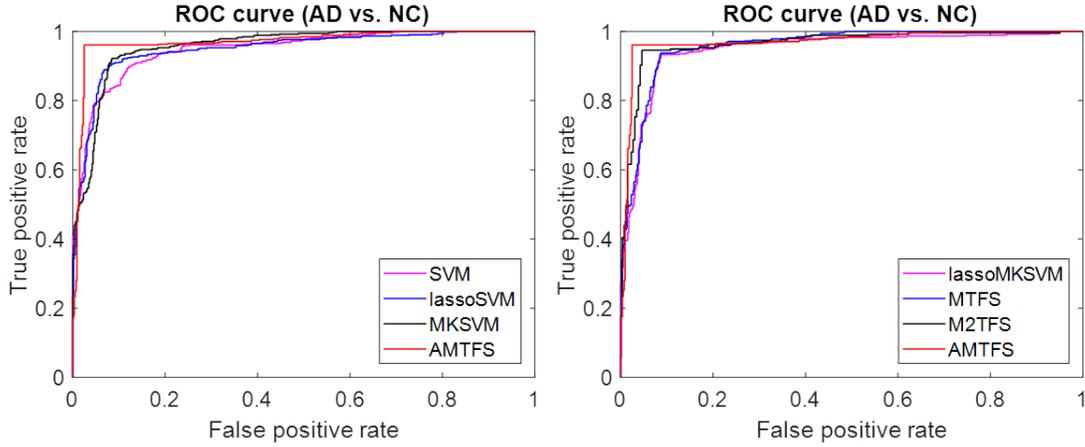

**Fig. 2** ROC curves of seven multi-modality based methods for classification of AD vs. NC

### 3.2.2 MCI vs. NC classification

Table 3 shows the performance of the seven classification methods in MCI vs. NC. As observed, the proposed method gets the best performance in accuracy, specificity, F1 Score and AUC, while M2TFS achieves the best sensitivity of 86.73%. Nevertheless, ASMFS is only 0.75% lower than M2TFS.

**Table 3** Comparison of performance of different methods for MCI vs. NC classification

| Method | Accuracy (%) | Sensitivity (%) | Specificity (%) | F1 Score | AUC |
| --- | --- | --- | --- | --- | --- |
| SVM | 70.62±0.1035 | 84.03±0.1176 | 45.20±0.2111 | 81.04±0.0599 | 0.7463±0.0013 |
| lassoSVM | 73.40±0.1167 | 81.62±0.1358 | 58.00±0.2141 | 79.78±0.0960 | 0.7852±0.0013 |
| MKSVM | 73.17±0.0983 | 80.69±0.1141 | 59.00±0.2189 | 79.62±0.0762 | 0.7276±0.0014 |
| lassoMKSVM | 74.19±0.0894 | 86.57±0.1098 | 50.70±0.2703 | 81.44±0.0647 | 0.7539±0.0012 |
| MTFS | 74.86±0.0911 | 82.19±0.1135 | 61.07±0.2066 | 80.91±0.0716 | 0.7296±0.0014 |
| M2TFS | 78.97±0.0766 | **86.73±0.1070** | 64.53±0.2515 | 84.35±0.0561 | 0.7526±0.0014 |
| ASMFS | **80.73±0.0950** | 85.98±0.1081 | **70.90±0.2135** | **85.30±0.0738** | **0.7875±0.0014** |

It should be pointed out that the accuracy of M2TFS and ASMFS is much higher than other methods. This is probably because the selected structural features play a pivotal role in the improvement of classification performance.

In addition, the results listed in Table 3 are generally lower than those in AD vs. NC classification. It is because the changes occurring in the brain of patients with MCI are less than those of patients with Alzheimer's disease. For instance, people with mild cognitive impairment have much less contraction in the hippocampus than people with AD. Hence, it is more difficult to identify patients with MCI than patients with AD. Figure 3 plots the corresponding ROC curves which reflect the classifier performance of the above algorithms.

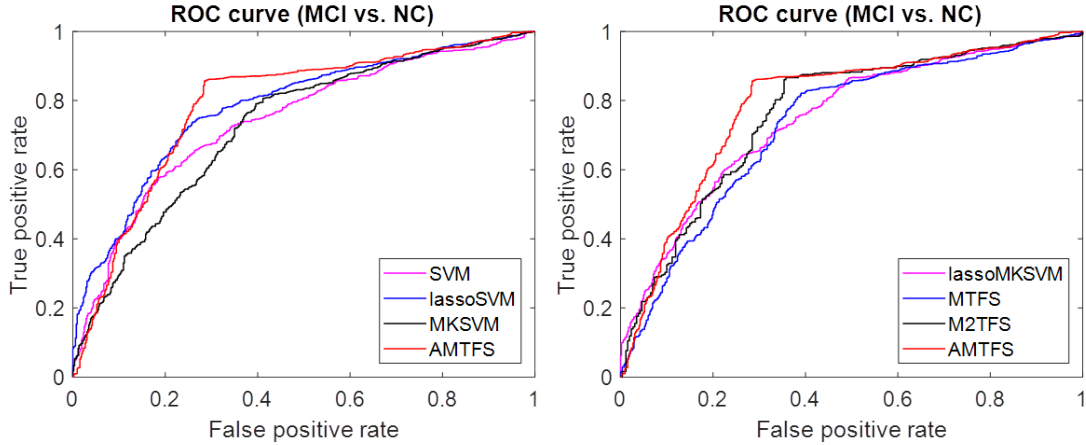
**Fig. 3** ROC curves of seven multi-modality based methods for classification of MCI vs. NC

### 3.2.3 MCI-C vs. MCI-NC classification

The classification results for MCI-C vs. MCI-NC are shown in Table 4. As can be seen from Table 4, our proposed method achieves a classification accuracy of 69.41 %, sensitivity of 65.3 % and F1 Score of 63.98 while the best sensitivity and F1 Score of the methods in comparison is only 54.5 % and 55.84 respectively, which is obtained by M2TFS. Besides, ASMFS achieves a specificity of 72.83% and AUC of 0.6534.

**Table 4** Comparison of performance of different methods for MCI-c vs. MCI-NC classification

| Method | Accuracy (%) | Sensitivity (%) | Specificity (%) | F1 Score | AUC |
| --- | --- | --- | --- | --- | --- |
| SVM | 56.45±0.1338 | 31.55±0.2126 | 75.90±0.2024 | 36.21±0.2195 | 0.6341±0.0017 |
| lassoSVM | 58.76±0.1394 | 48.75±0.2422 | 66.43±0.2127 | 48.69±0.1972 | 0.5830±0.0017 |
| MKSVM | 58.80±0.1206 | 54.45±0.2293 | 62.43±0.2202 | 51.74±0.1625 | 0.5753±0.0017 |
| lassoMKSVM | 61.73±0.1369 | 51.10±0.2469 | 70.23±0.2109 | 51.67±0.2032 | 0.6086±0.0018 |
| MTFS | 63.52±0.1220 | 59.65±0.2514 | 66.70±0.2108 | 56.63±0.1762 | 0.5894±0.0017 |
| M2TFS | 67.53±0.1059 | 54.50±0.2629 | **77.47±0.1873** | 55.84±0.2182 | **0.6647±0.0017** |
| ASMFS | **69.41±0.1194** | **65.30±0.2151** | 72.83±0.1811 | **63.98±0.1485** | 0.6534±0.0017 |

Similar to the results in MCI vs. NC, the accuracy of M2TFS and ASMFS is much higher than that of other methods which well demonstrates the importance of structural information. In these experiments, ASMFS achieves the highest accuracy owing to the adaptive similarity learning which optimizes the similarity between subjects and then enhances the discrimination of selected features.

It can also be found that the accuracy of the classification task is generally low. It is because that the differences between MCI-C and MCI-NC are small and the early symptoms of MCI are also similar for MCI converters and MCI non-converters. Besides, baseline MRI and PET data is utilized in our experiments and the converters have to be diagnosed in 18 months which also increases the difficulty for classification. However, as can be seen from Figure 4, the proposed method is still overall the best one.

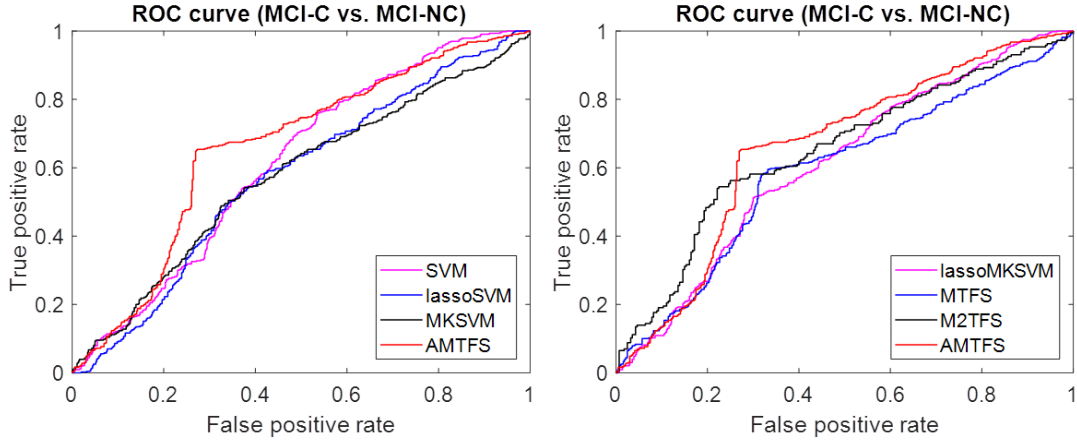

**Fig. 4** ROC curves of seven multi-modality based methods for classification of MCI-C vs. MCI-NC

### 3.3 Feature selection results

The most discriminative brain regions are defined as those that are ranked by the regression coefficient $W$. Figure 5, 6 and 7 and Table 5, 6 and 7 show the top 10 selected brain regions in the classification of AD vs. NC, MCI vs. NC and MCI-C vs. MCI-NC, respectively. For AD vs. NC classification, brain regions such as hippocampus, precuneus, uncus and temporal gyrus are found sensitive to AD by our proposed method. The brain regions, for instance, hippocampus and amygdala are also selected in MCI vs. NC classification tasks. The studies in [46, 47] have shown that the hippocampus is responsible for short-term memory, and in the early stage of Alzheimer's disease also known as MCI, hippocampus begins to be destroyed directly resulting in the decline of short-term memory and disorientation. Amygdala is a part of brain who is responsible for managing basic emotions such as fear and anger. The damage for amygdala caused by MCI/AD can result in paranoia and anxiety. The selected regions by the proposed method are consistent with many studies in literature. Moreover, the proposed method can also help researchers to focus on the brain regions selected in this experiment but neglected before, so as to help the diagnosis of MCI/AD.

**Table 5** Top 10 ROIs selected by the proposed method for AD vs. NC

| Number of ROIs | Selected ROIs |
| --- | --- |
| 69 | hippocampal formation left |
| 41 | precuneus left |
| 80 | middle temporal gyrus right |
| 84 | inferior temporal gyrus right |
| 18 | angular gyrus right |
| 87 | angular gyrus left |
| 26 | precuneus right |
| 46 | uncus left |
| 83 | amygdala right |
| 90 | lateral occipitotemporal gyrus left |

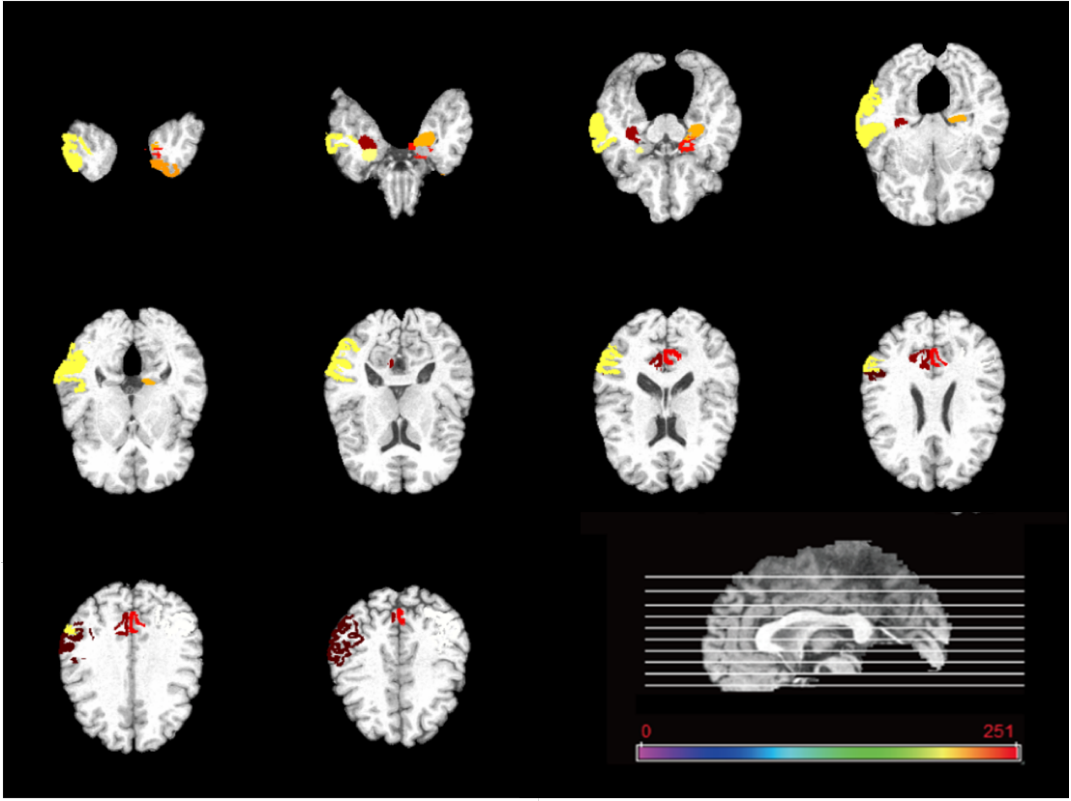

**Fig.5** Top 10 ROIs selected by the proposed method for AD vs. NC

**Table 6** Top 10 ROIs selected by the proposed method for MCI vs. NC

| Number of ROIs | Selected ROIs |
| --- | --- |
| 87 | angular gyrus left |
| 69 | hippocampal formation left |
| 64 | entorhinal cortex left |
| 40 | cuneus left |
| 83 | amygdala right |
| 41 | precuneus left |
| 63 | temporal pole left |
| 92 | occipital pole left |
| 30 | hippocampal formation right |
| 17 | parahippocampal gyrus left |

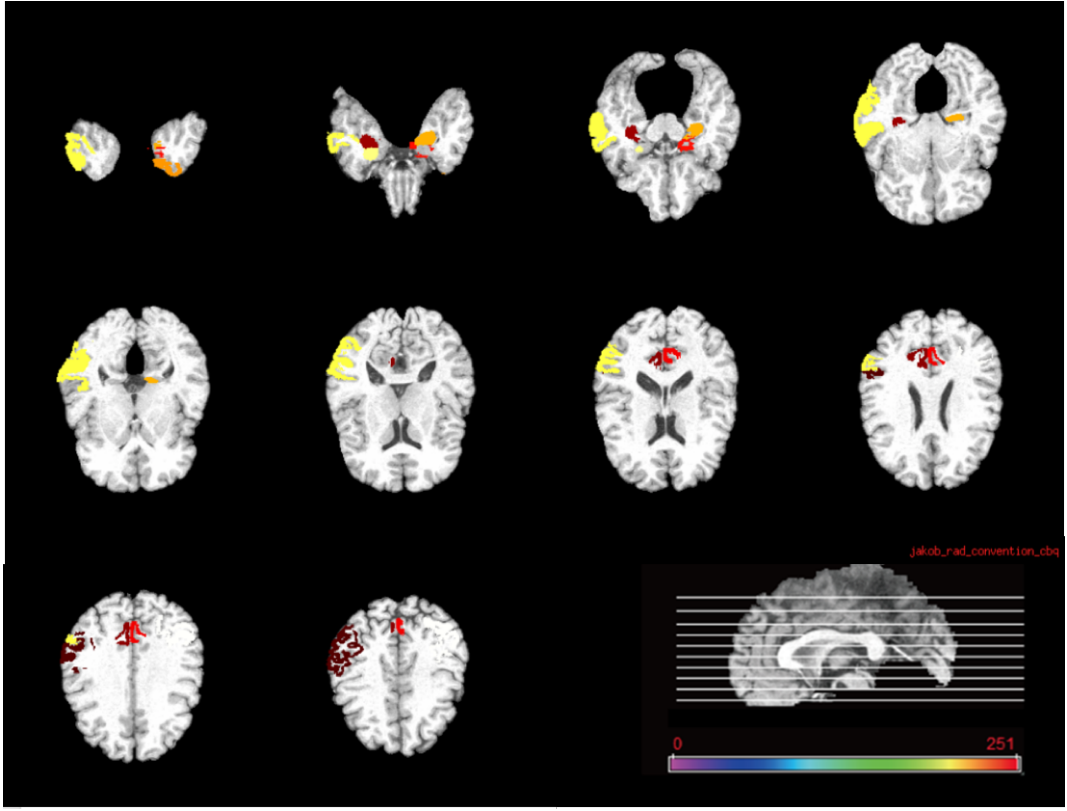

**Fig.6** Top 10 ROIs selected by the proposed method for MCI vs. NC

**Table 7** Top 10 ROIs selected by the proposed method for MCI-C vs. MCI-NC

| Number of ROIs | Selected ROIs |
|---|---|
| 41 | precuneus left |
| 61 | perirhinal cortex left |
| 35 | anterior limb of internal capsule left |
| 48 | middle temporal gyrus left |
| 10 | superior frontal gyrus right |
| 83 | amygdala right |
| 49 | lingual gyrus left |
| 86 | middle occipital gyrus left |
| 30 | hippocampal formation right |
| 24 | fornix left |

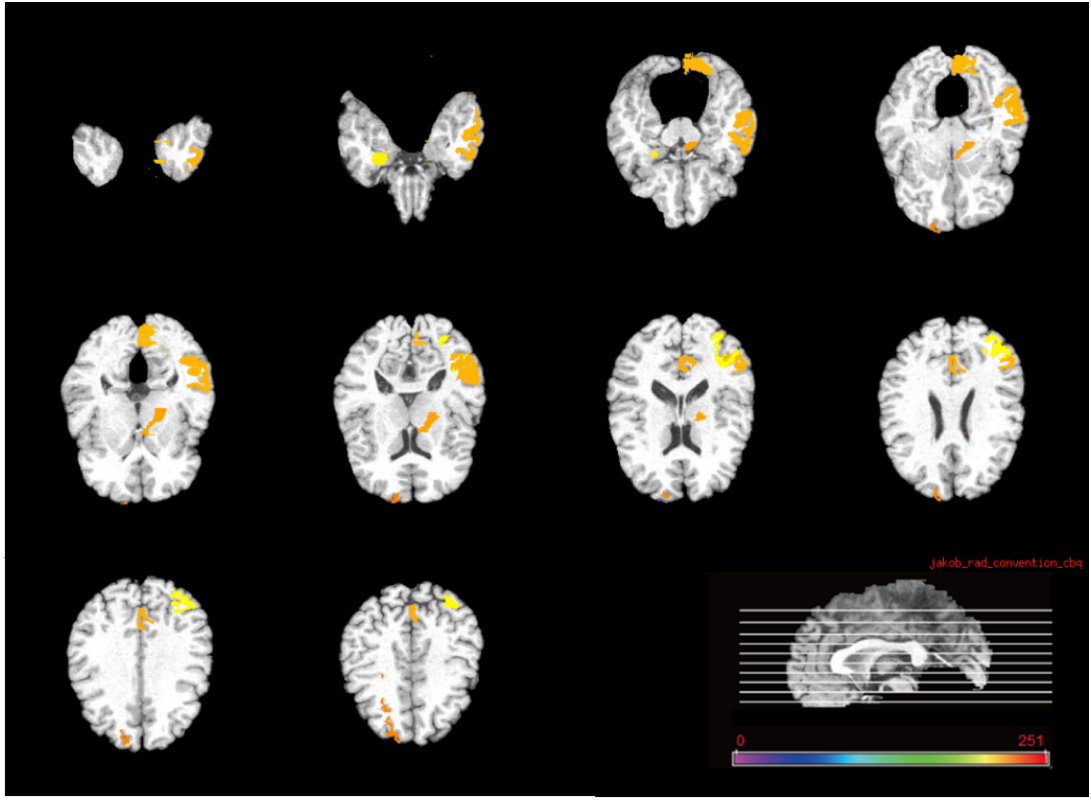

Fig.7 Top 10 ROIs selected by the proposed method for MCI-C vs. MCI-NC

### 3.4 Effect of hyper-parameters & Algorithm convergence

***Regularization parameters:*** In ASMFS, there are three hyper-parameters, i.e., $\lambda$, $\mu$ and $k$. Specifically, the adaptive similarity learning regularizer $\lambda$ and the group sparsity regularizer $\mu$ control the relative contribution of those regularization terms. $k$ is the number of neighbours in adaptive similarity learning. As aforementioned, the above parameters are determined by another 10-fold cross-validation. $\lambda$, $\mu$ and $k$ are searched in the range $\{0.1,5,20,60,100\}$, $\{0,5,10,15,20\}$ and $\{1,3,5,7,9\}$, respectively. It is worth mentioning that when $\mu = 0$, the group sparsity will cease to work. Then, all the features are retained for the subsequent classification which degenerates to the method of multi-modality classification proposed in [21].

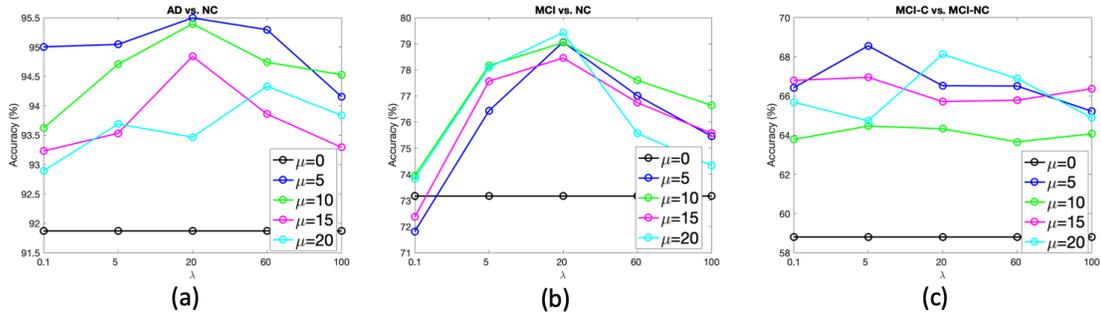

**Fig.8** The classification accuracy with regularization parameters $\lambda$ and $\mu$.

Figure 8 shows the classification results with regard to different values of $\lambda$ and $\mu$ when $k$ is fixed to 5. The X-axis indicates $\lambda$, Y-axis indicates classification accuracy and the curves with different colours represent different values of $\mu$ ranging from {0,5,10,15,20}, respectively. As observed, with the increase of $\lambda$ from 0.1 to 20, the curves corresponding to different values of $\mu$ are on rise, whereas the downtrend of accuracy is observed when $\lambda$ is larger than 20. Through analysis, we believe that $\lambda$ conducts a less effective guide when it is relatively small because $\lambda$ almost performs no contraint on the item of adaptive similarity learning.

Besides, as can be seen, when $\lambda$ is fixed, $\mu$ has a greater impact on classification accuracy than $\lambda$, which is because $\mu$ affects the sparsity of $W$ and determines the number of discriminative features. Also, as we can see from Figure 8 (a), when $\mu = 0$ which suggests that no feature is selected, the corresponding accuracy curve lies below other curves. The similar phenomenon can be seen from Figure 8 (b), (c) as well. Such result demonstrates the effectiveness of feature selection.

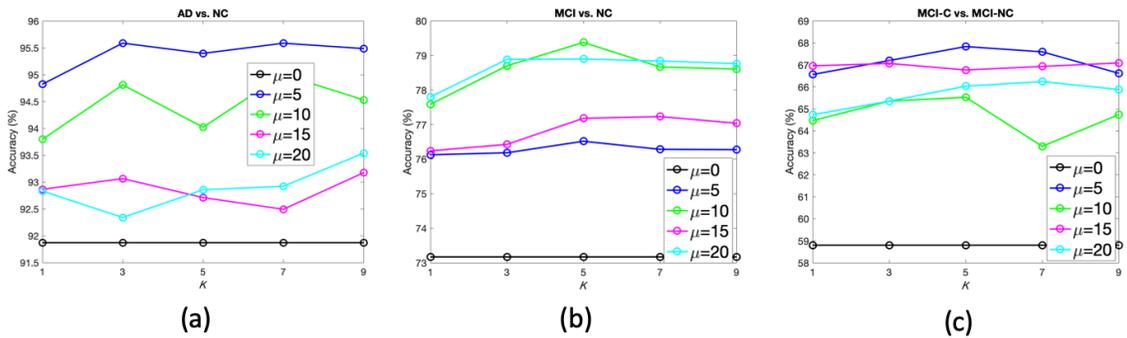

**Fig.9** The classification accuracy with regularization parameters $k$ and $\mu$

Figure 9 shows the classification results with different values of $k$ and $\mu$ when $\lambda$ is fixed to 20. As observed, the classification performance with feature selection ($\mu = 5,10,15,20$) is better than that without feature selection ($\mu = 0$). Most of the curves reach its peak when $k = 5$ but go down when $k$ is beyond 5. Such result suggests that maintaining the local manifold structure of data helps to select discriminative features. What's more, Figure 9 shows the similar profile of curves with different values of $\mu$ when compared with Figure 8.

As can be seen from the above experimental results, a set of proper parameters plays an essential role in the classification performance.

*Algorithm convergence:* We also investigate the convergency of the proposed method. As we can observe from Figure 10, the proposed optimization algorithm has a good convergence, since the algorithm has basically converged after about 10 iterations.

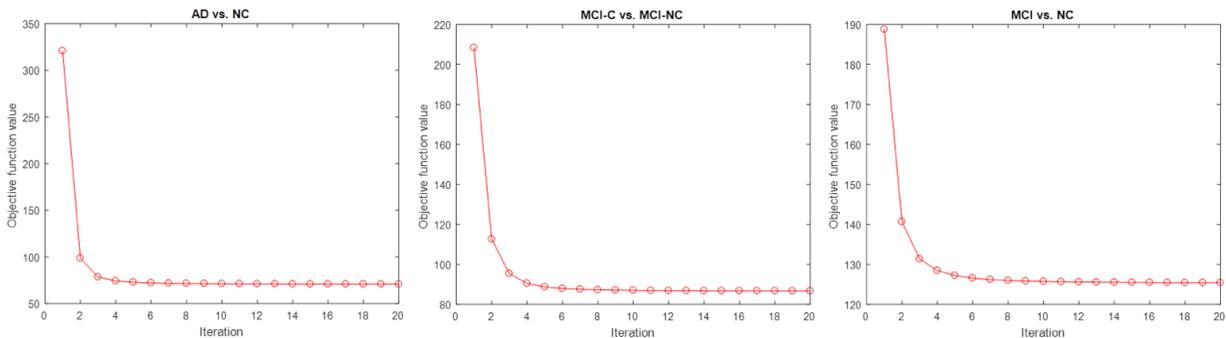

**Fig.10** Algorithm convergence of ATMFS

## 3.5 Comparison with the state-of-the-art methods

Furthermore, we compare the results achieved by our method with several recent state-of-the-art results reported in the literature studying the multi-modality method. We list the details of each method and the corresponding results in Table 8.

As can be seen from Table 8, Hinrichs et al. [55] used 48 AD subjects and 66 NC subjects, and obtained an accuracy of 87.6% by using two modalities (MRI+PET). Huang et al. [55] used 49 AD patients and 67 NC with MRI and PET modalities for AD classification, achieving an accuracy of 94.3%. In [22], Gray et al. used 37 AD patients, 75 MCI patients and 35 NC and reported classification accuracies of 89.0%, 74.6% and 58.0% for AD, MCI and MCI-converter classification, respectively, using four different modalities (MRI+PET+CSF+genetic). Jie et al. [26] achieved the accuracies of 95.03%, 79.27% and 68.94% for classification of AD/NC, MCI/ NC and MCI-C/MCI-NC, respectively. Liu et al. [25] obtained the accuracies of 94.37%, 78.80% and 67.83 % for AD, MCI and MCI-converter classifications, respectively. Zu et al. [54] achieved the accuracies of 95.95, 80.26 and 69.78%, respectively. The dataset used in [26], [25] and [54] is the same as this study. Table 8 indicates that our proposed method outperforms other methods, which further validates the efficacy of our proposed method for AD diagnosis.

**Table 8** Comparison of classification accuracy of different multi-modality methods

| Method | Subjects | Modalities | AD vs. NC | MCI vs. NC | MCI-C vs. MCI-NC |
| --- | --- | --- | --- | --- | --- |
| Hinrichs et al. [55] | 48 AD + 66 NC | MRI + PET | 87.6% | - | - |
| Huang et al. [56] | 49 AD + 67 NC | MRI + PET | 94.3% | - | - |
| Gray et al. [22] | 37 AD + 75 MCI + 35 NC | MRI + PET + CSF + genetic | 89.0% | 74.6% | 58.0% |
| Jie et al. [26] | 51 AD + 99 MCI + 52 NC | MRI + PET | 95.03% | 79.27% | 68.94% |
| Liu et al. [25] | 51 AD + 99 MCI + 52 NC | MRI + PET | 94.37% | 78.80% | 67.83% |
| Zu et al. [54] | 51 AD + 99 MCI + 52 NC | MRI + PET | 95.95% | 80.26% | **69.78%** |
| Proposed | 51 AD + 99 MCI + 52 NC | MRI + PET | **96.76%** | **80.73%** | 69.41% |

## 4. Discussion

Multi-modality learning, a recently developed technique in machine learning field which can jointly learn multiple modalities via a shared representation, has been successfully used across all applications of machine learning, from natural language processing [48] and speech recognition [49] to computer vision [50] and drug discovery [51]. Recently, multi-modality learning has been introduced into medical imaging field. However, the problem of small number of subjects and high feature dimensions limits further performance improvement of the multimodal classification methods. Our work aims to provide a novel multi-modality feature selection method which not only reduces irrelevant and redundant features but also considers the local similarity across different imaging modalities. Although the idea of jointly selecting features from multi-modality neuroimaging data has been seen in previous study [44, 45, 54], these methods do not consider the potential relationship across different modalities. Besides, underlying data structure in the low-dimensional space may not be revealed in these methods since the neighbors and similarity of the original high-dimensional data are obtained separately from each individual modality.

In this paper, we apply an adaptive similarity learning method to address the above issues. The similarity measured from single and multi-modality data is learned with the change of low-dimensional representation after feature selection. As can be observed from the experimental results in section 3.2, our multi-modality feature learning method which adopts adaptive similarity learning method shows better performance than those with fixed similarity based methods, thus the superiority of adaptive similarity learning for feature selection is fully demonstrated.

Besides, we keep the local neighborhood structure during similarity learning by updating *K* local neighborhood similarities of each subject. This idea was inspired by [54] which suggests that each data point and its neighbors lie on or near a locally linear patch of the manifold. Thus, one could depict the local geometry of linear patches by linear coefficients that reconstruct each data point from its neighbors. Experiments conducted in section 3.4 corroborate that keeping the local manifold structure of data is beneficial to feature selection.

Several limitations should be further considered in future studies. First, in this paper, we only considered two-class classification problem (i.e., AD vs. NC, MCI vs. NC, and MCI-C vs. MCI-NC), and do not test the ability of our proposed method for the multi-class classification of AD, MCI and NC. Although multi-class classification is more challenging than binary classification, it is crucial to diagnose the different stages of dementia which is helpful for doctors to suit the remedy to the case. Second, the proposed method requires the same number of features from different modalities. However, other modalities in the ADNI database which may also contain important pathological information have different feature numbers. Consequently, modalities like CSF and genetic data cannot play their roles in classification. Finally, longitudinal image data may carry important information for prediction, while the proposed method in this paper does not consider the data at multiple time points. Thus, for future work, we plan to extend ASMFS to a multi-class classification method which can process more modalities with different number of features. Besides, it is also interesting to investigate other strategies to utilize the hidden information contained in longitudinal image data.

## 5. Conclusion

This paper proposes a novel multi-modality method where feature selection and similarity learning are learned jointly and we call this method Adaptive-Similarity-based Multi-modality Feature Selection (ASMFS). By introducing the adaptive similarity learning mechanism into the multi-modality learning framework, the proposed method can not only fully explore the relationships across both modalities and subjects through mining and fusing discriminative features from multi-modality data for AD/MCI classification, but also capture the intrinsic data structure of different modality data in the low-dimensional space. Experimental results on the ADNI database demonstrate that our proposed method outperforms the state-of-the-art methods for multimodal classification of AD/MCI.

**Acknowledgements**

This work was supported by National Natural Science Foundation of China (NSFC61701324).

# Conflict of interest

The authors declare no conflict of interest.